\documentclass[letterpaper, 10 pt, conference]{class/ieeeconf}  % Comment this line out
\IEEEoverridecommandlockouts                              % This command is only
\newcommand{\red}[1]{\textcolor{red}{#1}}
\usepackage{amsmath}
\usepackage{amssymb}
\usepackage{latexsym}
\usepackage{url}
\usepackage{cite}
\usepackage{relsize}
\usepackage{multirow}
\usepackage{afterpage}
\usepackage{ifthen}
\usepackage{graphicx}
\usepackage{algpseudocode,algorithm,algorithmicx}
\usepackage{subfigure}
\usepackage{epstopdf}
\usepackage{stackengine}
\usepackage{gensymb}
\usepackage[bookmarks=false, hidelinks]{hyperref} 

\usepackage{booktabs}
\usepackage{float}
\usepackage{tablefootnote}
\usepackage{makecell}
\usepackage{hhline}
\usepackage{courier}
\usepackage{lipsum}
\usepackage{booktabs}
\usepackage{lettrine}
\usepackage{tabu}
\usepackage{nicematrix,tikz}
\usepackage{arydshln}
\usepackage{tikz}
\usepackage{adjustbox}
\usepackage{pgfplots}
\usepgfplotslibrary{groupplots}
\usetikzlibrary{pgfplots.groupplots}
\usetikzlibrary{plotmarks}
\usetikzlibrary{calc}
\usepgfplotslibrary{external}
\pgfplotsset{compat=newest}

\usetikzlibrary{matrix}
\pgfplotsset{grid style={dashed,gray}}

\hypersetup{
    colorlinks=true,
    linkcolor=red,
    filecolor=magenta,      
    urlcolor=blue,
    pdfstartview={FitH}    }

\usepackage{amssymb}% http://ctan.org/pkg/amssymb
\usepackage{pifont}% http://ctan.org/pkg/pifont
\newcommand{\xmark}{\ding{55}}%

\title{\LARGE \bf CathSim: An Open-source Simulator for Endovascular Intervention}

\author{Tudor Jianu$^1$, Baoru Huang$^2$, Minh Nhat Vu$^3$, Mohamed E. M. K. Abdelaziz$^2$, Sebastiano Fichera$^4$,\\ Chun-Yi Lee$^5$, Pierre Berthet-Rayne$^6$, Ferdinando Rodriguez y Baena$^7$, Anh Nguyen$^1$ 
\thanks{$^1$Department of Computer Science, University of Liverpool, UK}
\thanks{$^2$The Hamlyn Centre for Robotic Surgery, Imperial College London, UK}
\thanks{$^3$Automation \& Control Institute (ACIN), TU Wien, Austria}
\thanks{$^4$Deparment of Mechanical, Materials \& Aerospace Engineering, University of Liverpool, UK}
\thanks{$^5$Department of Computer Science, National Tsing Hua University, Taiwan}
\thanks{$^6$Honorary Fellow, University of Liverpool}
\thanks{$^7$Mechatronicsin Medicine Lab, Imperial College London, UK.}}
\begin{document}
% Macros

\newtheorem{problem}{Problem}
\newtheorem{lemma}{Lemma}
\newtheorem{theorem}[lemma]{Theorem}
\newtheorem{claim}{Claim}
\newtheorem{corollary}[lemma]{Corollary}
\newtheorem{definition}[lemma]{Definition}
\newtheorem{proposition}[lemma]{Proposition}
\newtheorem{remark}[lemma]{Remark}
\newenvironment{LabeledProof}[1]{\noindent{\it Proof of #1: }}{\qed}

\def\beq#1\eeq{\begin{equation}#1\end{equation}}
\def\bea#1\eea{\begin{align}#1\end{align}}
\def\beg#1\eeg{\begin{gather}#1\end{gather}}
\def\beqs#1\eeqs{\begin{equation*}#1\end{equation*}}
\def\beas#1\eeas{\begin{align*}#1\end{align*}}
\def\begs#1\eegs{\begin{gather*}#1\end{gather*}}

\newcommand{\poly}{\mathrm{poly}}
\newcommand{\eps}{\epsilon}
\newcommand{\e}{\epsilon}
\newcommand{\polylog}{\mathrm{polylog}}
\newcommand{\rob}[1]{\left( #1 \right)} %Round Brackets
\newcommand{\sqb}[1]{\left[ #1 \right]} %square Brackets
\newcommand{\cub}[1]{\left\{ #1 \right\} } %curly brackets
\newcommand{\rb}[1]{\left( #1 \right)} %Round
\newcommand{\abs}[1]{\left| #1 \right|} %| |
\newcommand{\zo}{\{0, 1\}}
\newcommand{\zonzo}{\zo^n \to \zo}
\newcommand{\zokzo}{\zo^k \to \zo}
\newcommand{\zot}{\{0,1,2\}}
\newcommand{\en}[1]{\marginpar{\textbf{#1}}}
\newcommand{\efn}[1]{\footnote{\textbf{#1}}}
\newcommand{\vecbm}[1]{\boldmath{#1}} %more general (handles greek letters)
\newcommand{\uvec}[1]{\hat{\vec{#1}}}
\newcommand{\thv}{\vecbm{\theta}}
\newcommand{\junk}[1]{}
\newcommand{\var}{\mathop{\mathrm{var}}}
\newcommand{\rank}{\mathop{\mathrm{rank}}}
\newcommand{\diag}{\mathop{\mathrm{diag}}}
\newcommand{\tr}{\mathop{\mathrm{tr}}}
\newcommand{\acos}{\mathop{\mathrm{acos}}}
\newcommand{\atantwo}{\mathop{\mathrm{atan2}}}
\newcommand{\SVD}{\mathop{\mathrm{SVD}}}
\newcommand{\quadf}{\mathop{\mathrm{q}}}
\newcommand{\linterp}{\mathop{\mathrm{l}}}
\newcommand{\sgn}{\mathop{\mathrm{sign}}}
\newcommand{\sym}{\mathop{\mathrm{sym}}}
\newcommand{\avg}{\mathop{\mathrm{avg}}}
\newcommand{\mean}{\mathop{\mathrm{mean}}}
\newcommand{\erf}{\mathop{\mathrm{erf}}}
\newcommand{\grad}{\nabla}
\newcommand{\R}{\mathbb{R}}
\newcommand{\defeq}{\triangleq}
\newcommand{\dims}[2]{[#1\!\times\!#2]}
\newcommand{\sdims}[2]{\mathsmaller{#1\!\times\!#2}}
\newcommand{\udims}[3]{#1}
\newcommand{\udimst}[4]{#1}
\newcommand{\com}[1]{\rhd\text{\emph{#1}}}
\newcommand{\ind}{\hspace{1em}}
\newcommand{\argmin}[1]{\underset{#1}{\operatorname{argmin}}}
\newcommand{\floor}[1]{\left\lfloor{#1}\right\rfloor}
\newcommand{\step}[1]{\vspace{0.5em}\noindent{#1}}
\newcommand{\quat}[1]{\ensuremath{\mathring{\mathbf{#1}}}}
\newcommand{\norm}[1]{\left\lVert#1\right\rVert}
\newcommand{\ignore}[1]{}
\newcommand{\specialcell}[2][c]{\begin{tabular}[#1]{@{}c@{}}#2\end{tabular}}
\newcommand*\Let[2]{\State #1 $\gets$ #2}
\newcommand{\algorithmicbreak}{\textbf{break}}
\newcommand{\Break}{\State \algorithmicbreak}

\renewcommand{\vec}[1]{\mathbf{#1}} %looks better

\algdef{S}[FOR]{ForEach}[1]{\algorithmicforeach\ #1\ \algorithmicdo}
\algnewcommand\algorithmicforeach{\textbf{for each}}
\algrenewcommand\algorithmicrequire{\textbf{Require:}}
\algrenewcommand\algorithmicensure{\textbf{Ensure:}}
\algnewcommand\algorithmicinput{\textbf{Input:}}
\algnewcommand\INPUT{\item[\algorithmicinput]}
\algnewcommand\algorithmicoutput{\textbf{Output:}}
\algnewcommand\OUTPUT{\item[\algorithmicoutput]}

% Tudor

\maketitle
 \thispagestyle{empty}
\pagestyle{empty}

\def\etal{\emph{et al.}}

%%%%%%%%%%%%%%%%%%%%%%%%%%%%%%%%%%%%%%%%%%%%%%%%%%%%%%%%%%%%%%%%%%%%%%%%%%%%%%%%
\begin{abstract}

% \textbf{Three things when writing the abstract: 1. What is the problem we want to solve. 2. how our method solve it. 3. Our results. Focus on OUR method, our strength ..., not others.}
% Problem | The lack of a benchmarking environment for endovascular procedures
% The autonomy of endovascular robotic systems is increasing where robotic systems such as Corindus gained FDA clearance for the first automated task within CorPath system. However, the process of training an autonomous system in reality suffers from costs in terms of time and can inflict damage on the hardware. 

% Autonomous robots in endovascular procedures have the potential to safely and reliably navigate the cardiovascular systems whilst reducing the need for human mistakes. However, there are numerous challenges associated with the process of training such robots, such as the time component implied by the sample inefficiency of the state-of-the art algorithms as well as the cost implied by the chance of material damage.
% We propose a simulation environment that allows rapid training and benchmarking of reinforcement learning algorithms within the task of catheter navigation for endovascular procedures. The simulated environment is built considering simplicity, speed and modularity by using the OpenAI Gym framework in conjunction with MuJoCo's physics engine.

% Problem
Autonomous robots in endovascular operations have the potential to navigate circulatory systems safely and reliably while decreasing the susceptibility to human errors. However, there are numerous challenges involved with the process of training such robots, such as long training duration and safety issues arising from the interaction between the catheter and the aorta. Recently, endovascular simulators have been employed for medical training but generally do not conform to autonomous catheterization. Furthermore, most current simulators are closed-source, which hinders the collaborative development of safe and reliable autonomous systems. In this work, we introduce CathSim, an open-source simulation environment that accelerates the development of machine learning algorithms for autonomous endovascular navigation. We first simulate the high-fidelity catheter and aorta with a state-of-the-art endovascular robot. We then provide the capability of real-time force sensing between the catheter and the aorta in simulation. Furthermore, we validate our simulator by conducting two different catheterization tasks using two popular reinforcement learning algorithms. The experimental results show that our open-source simulator can mimic the behaviour of real-world endovascular robots and facilitate the development of different autonomous catheterization tasks. Our simulator is publicly available at \href{https://github.com/robotvisionlabs/cathsim}{https://github.com/robotvisionlabs/cathsim}.

%  , the brachiocephalic artery and the left common carotid artery. Furthermore, we provide two different types of environments (i.e. aortic arches) as well as three observation spaces (i.e. internal state, top-view image and sequential images).
% % Evaluation
% We evaluate our environment considering the force resulted by the interaction between the catheter and the aorta and provide initial results of Reinforcement Learning training by using the

% \lipsum[1]
\end{abstract}

%%%%%%%%%%%%%%%%%%%%%%%%%%%%%%%%%%%%%%%%%%%%%%%%%%%%%%%%%%%%%%%%%%%%%%%%%%%%%%%%
\section{INTRODUCTION} \label{Sec:Intro}

\lettrine[lines=2]{\textbf{E}}{ndovascular} intervention has been continuously evolving since the traditional approach of direct, open-cut surgery. It involves the use of a small incision that allows surgical equipment (such as catheters and guidewires) to be manoeuvred within the vasculature.  This type of minimally invasive surgery (MIS) provides numerous advantages where the patient benefits from reduced blood loss, shorter recovery time, lower postoperative pain, and diminished inflammatory response compared to the traditional approaches~\cite{wamala2017use}. In typical clinical conditions, the catheter and guidewire are navigated to the diagnosis zone through the use of fluoroscopy, a medical visualization procedure that obtains real-time X-Ray images from the operating theatre. Despite the relative advantages, endovascular intervention still presents some drawbacks such as lack of sensory feedback, surgeon exposure to radiation, and the need for highly dexterous manipulation~\cite{omisore2020review}.

To reduce the continuous risk of radiation imposed on the surgeon throughout the fluoroscopic procedure, many robotic systems with leader-follower teleoperation architecture have been proposed~\cite{kundrat2021mr, pereira2020first, burgner2015continuum}. The surgeon actuates the leader device, from which, the information is mapped to the follower robot that executes the related action. The use of leader-follower robots allows the surgeon to perform the procedure remotely from a safe, radiation-free zone. Recent work has further focused on creating Magnetic Resonance (MR) safe robotic platforms\cite{kundrat2021mr}, which eliminates the ionizing radiation exposure whilst allowing the soft tissue, such as the vasculature, to be visualized~\cite{heidt2019real, abdelaziz2019toward}. Furthermore, in academic settings, the robotic system is developed to provide additional information to the surgeon through assistive features such as force/torque information~\cite{konstantinova2014implementation}, haptic feedback~\cite{molinero2019haptic}, and real-time segmentation and tracking~\cite{nguyen2020end}. Nonetheless, the procedure continues to be manually conducted by the surgeon due to the inherent technical, regulatory, ethical, and legal challenges~\cite{dupont2021decade}.

\begin{figure}[t]
	\centering
	\includegraphics[scale=0.20]{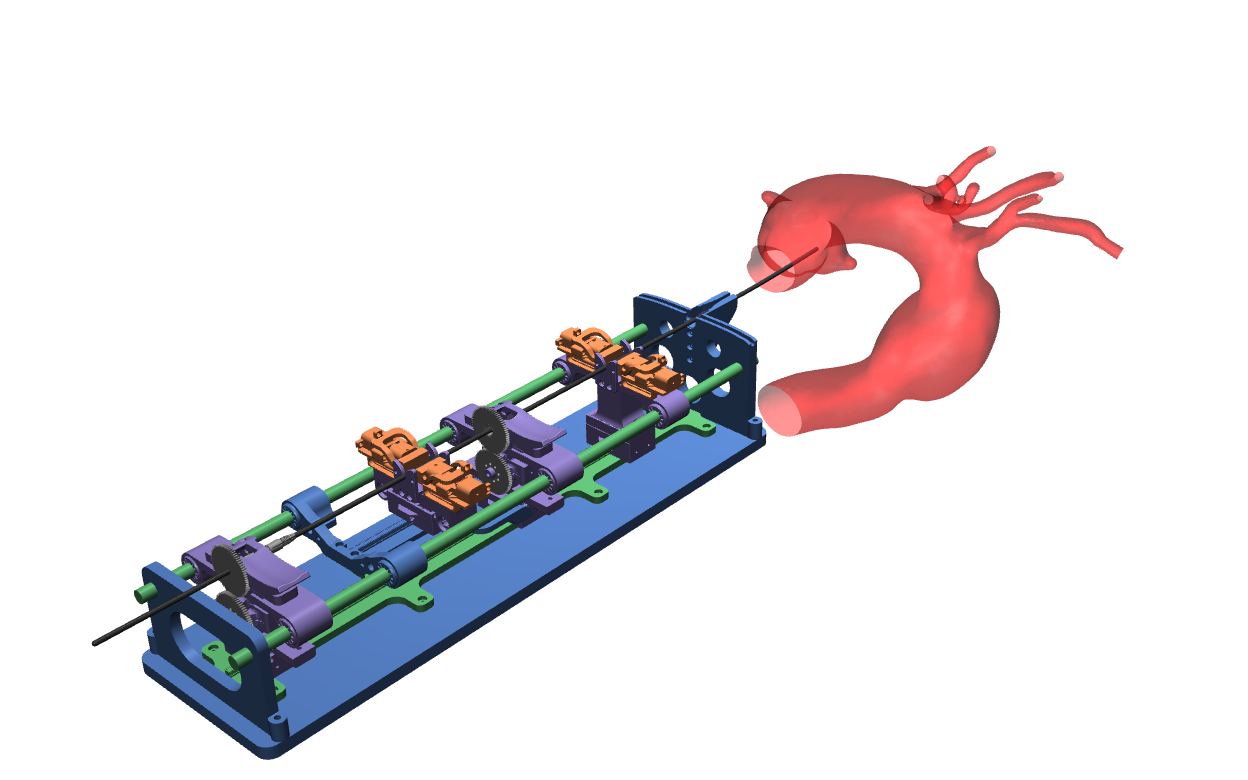}
    \vspace{1ex}
	\caption{An overview of our \textbf{CathSim} - an open-sourced simulator for autonomous cannulation. The figure shows the follower robot navigating the catheter through an aortic arch.
	}
	\vspace{0.3cm}
	\label{fig:overview}
\end{figure}

 \begin{table*}[t!]
\caption{Endovascular Simulator Environments Comparison}
\def\arraystretch{1.2}
\begin{center}
\begin{tabular}{l c c c c c c c }
\toprule
 Simulator & Physics Engine & Catheter  & Guidewire & Aorta %& Force Sensing %& Open-source  
 \\
\midrule
Molinero~\etal~\cite{molinero2019haptic}         & Unity Physics~\cite{juliani2018unity} & Discretized                                       & \xmark & 3D artery %& Vision-Based %& \xmark     
\\
Karstensen~\etal~\cite{karstensen2020autonomous} & SOFA~\cite{faure:hal-00681539}        & Timoshenko Beam theory~\cite{davis1972timoshenko} & \xmark & 2D vessel %& \xmark       %& \xmark     
\\
Behr~\etal~\cite{behr2019deep}                   & SOFA~\cite{faure:hal-00681539}        & Timoshenko Beam theory~\cite{davis1972timoshenko}  & \xmark & 2D vessel %& \xmark       %& \xmark     
\\
Omisore~\etal~\cite{omisore2021novel}            & CopelliaSim~\cite{rohmer2013v}        & \xmark                                           & Discretized & 3D vessel %& \xmark       %& \xmark     
\\
Schegg~\etal~\cite{schegg2022automated}          & SOFA~\cite{faure:hal-00681539}        & \xmark  & Timoshenko Beam theory~\cite{davis1972timoshenko}  & 3D artery %& \xmark       %& \xmark     
\\
\hline
\textbf{CathSim} (ours)                                   & MuJoCo~\cite{mujoco}                  & Discretized    & \xmark & 3D artery                                    %& \xmark & \checkmark   %& \checkmark 
\\
\bottomrule
%\multicolumn{6}{l}{\textit{Note: Whilst Timoshenko Beam catheter plugin is open-sourced, the whole simulation environment is not.}}
\end{tabular}
\label{tab:simulators}
\end{center}
\end{table*}

Whilst recent robotic platforms for endovascular intervention demonstrate their assistive potential in the successful completion of the procedure, they share two problems:
 \textit{i)} the lack of autonomy~\cite{attanasio2021autonomy} and
 \textit{ii)} increased duration of robotic procedure compared to its non-robotic counterpart~\cite{chi2020collaborative}. Firstly, the surgeon operates within the tridimensional space of the vasculature whilst relying on the information provided by two-dimensional fluoroscopic images and haptic feedback. Additionally, the surgeon has to avoid inflicting extensive damage to the vasculature system, thus operating under mentally strenuous conditions which can be diminished through the automation of the procedure. % Furthermore, the robotic procedure duration exposes the patient to higher risks, although it inflicts less force on the vasculature~\cite{kundrat2021mr}. 
A more autonomous surgery would ideally inflict little damage whilst operating in a timely manner. However, this is not a trivial task in practice as it requires a complex vision, learning, and control system that can guarantee the safety of the procedure~\cite{attanasio2021autonomy}.
 
Recent developments in machine learning promise a greater degree of autonomy in many robotic systems. Such systems leverage deep learning architectures such as convolutional neural networks~\cite{marban2019recurrent,gao2018learning}, long-short-term memory~\cite{aviles2016towards, aviles2016deep, marban2018estimation}, and generative adversarial imitation learning~\cite{nguyen2020end} to facilitate force estimation and catheter segmentation. Whilst those systems confer a lower level autonomy through the use of robotic assistive features, the higher level autonomy is left unaddressed. However, the navigation task has been addressed through many works~\cite{karstensen2020autonomous,omisore2021novel, chi2020collaborative}. The environment employed by those works makes tradeoffs between the use of physical and virtual environments, and they generally rely on closed-source environments which cannot be replicated by fellow researchers. Furthermore,  albeit the use of simulated environments, they generally do not adhere to the Reinforcement Learning (RL) paradigm.

In this work, our goal is to provide a new minimally invasive surgery environment for endovascular procedures. We aim at facilitating the development of endovascular autonomy through the provision of a standardized environment that confers familiarity to the machine learning community. As such, we propose \textbf{CathSim}, a real-time simulation environment for autonomous cannulation based on MuJoCo~\cite{mujoco}. We choose MuJoCo as the base simulator as it is a real-time and accurate physics engine that facilitates optimal control applications, hence well-suited for our task. Fig.~\ref{fig:overview} shows the overview of our simulator. We summarize our contributions and the potential usage of our simulator as follows:

\begin{enumerate}
    \item We propose \textbf{CathSim}, a new open-source simulation environment for endovascular procedures.
    \item We implement the baseline and provide the benchmark for autonomous cannulation tasks in our simulator using two popular RL algorithms.
    \item With real-time force sensing and high-fidelity visualization, our open-source simulator can be used in various tasks such as medical training or Augmented Reality (AR) and Virtual Reality (VR) applications.  
\end{enumerate}

\section{Related Work} \label{Sec:rw}

\textbf{Simulation Environments.} Research within the simulation of environments for minimally invasive surgery divides the simulation level into four distinct categories: synthetic, animal, virtual reality and human cadaver\cite{nesbitt2016role} where each possesses its own limitations and advantages~\cite{wei2012near,dequidt2009towards,talbot2014interactive}. They mostly concentrate on the skill development of the trainee~\cite{sinceri2015basic, nesbitt2016role}, or the development of assistive features such as haptic feedback \cite{molinero2019haptic}, or rely on physical materials which do not offer an RL-compliant environment. Recent research has been undertaken using synthetic simulators such as a high fidelity phantom~\cite{chi2020collaborative} by means of imitation learning, whereas other research used the SOFA~\cite{faure:hal-00681539} simulation environment~\cite{lillicrap2015continuous} and tested on a bi-dimensional synthetic phantom. Whilst the literature is advancing, the physical or closed-source nature of such simulators hinders the ability for collaborative improvement.

\textbf{Autonomous Catheterization.} Machine learning has facilitated the shift from assistive features to semi-autonomous navigation in autonomous catheterization~\cite{yang2017medical}. Our research focuses on deep reinforcement learning (RL) for its role in complex decision-making, as proven in fields like autonomous driving. Many studies utilize RL, particularly using images from fluoroscopy~\cite{karstensen2020autonomous,kweon2021deep,molinero2019haptic,behr2019deep,omisore2021novel}. While alternative methods, like the Dijkstra algorithm or breadth-first search exist~\cite{cho2021image,schegg2022automated}, model-free RL is suitable for managing the uncertainty and complexity of achieving higher autonomy. However, most research is still in the early stages of autonomy~\cite{yang2017medical}, making comprehensive autonomous navigation of the vascular system a challenging yet prospective target for reinforcement learning.

In Table~\ref{tab:simulators}, we show a detailed comparison of current learning-based works that make use of environments based on a variety of physics engines. A limitation of such environments is that they are not publicly available, which hinders the reproducibility. \textbf{CathSim}, in contrast to other simulators, provides an open-source environment that is well-suited for training autonomous agents using different machine-learning approaches. Based on MuJoCo's~\cite{mujoco} framework, our simulator offers an advanced simulation environment for real-time applications. Furthermore, our simulator provides real-time force sensing capability and high-fidelity realistic visualization of the aorta, catheter, and endovascular robots. In practice, \textbf{CathSim} can be used to train RL agents or serve as a practice platform for healthcare professionals. % Based on its flexible design, we can also easily extend CathSim to other environments such as Augmented Reality (AR) and Virtual Reality (VR) applications.

%%%%%%%%%%%%%%%%%%%%%%%%%%%%%%%%%%%%%%%%%%%%%%%%%%%%%%%%%%%%%%%%%%%%%%%%%%%%

% \subsubsection{Endovascular Robotic Platforms} Recently, many robotic platforms have been proposed for endovascular intervention. Examples of such platforms are Magellan (AurisHealth, Redwood City, CA, USA), the R-oneTM robot (Robocath, Rouen, France), the Amigo platform (Catheter Precision, Mt. Olive, NJ, USA), and the CorPath GRX platform (Corindus, A Siemens Healthineers Company, Waltham, MA, USA). Those platforms are designed for teleoperation and typically employ a leader-follower architecture~\cite{howe1999robotics} which allows the surgeon to be situated outside of the operating room, thus diminishing the risk of repeated exposure to ionising radiation that the surgeon is facing. The surgeon manipulates the leader robot which serves as an input to the follower robot. This information is then mapped to the end effector and visual-haptic feedback is provided back to the surgeon. Such platforms are designed for single-port access and due to the dimensionality imposed by the confined spaces they operate in, the actuation is fairly limited. Furthermore, the space-compliant modelling of continuum catheters implies great challenges in terms of actuation control, contact handling and sensorial perception~\cite{simaan2018medical}. 
\section{The \textbf{CathSim} Simulator}
Our \textbf{CathSim} environment has three components: \textit{i)} the follower robotic model for endovascular procedures~\cite{abdelaziz2019toward}, \textit{ii)} the aortic arch phantoms, and \textit{iii)} the catheter. Our simulator enables real-time simulation and supports the training of state-of-the-art learning algorithms.

\subsection{Robot Simulation}
In this work, we aim at transferring CathBot~\cite{kundrat2021mr} to simulation with the purpose of autonomous agents training. We choose CathBot as it is a state-of-the-art robot and is not bounded by a commercial licence. The design of CathBot follows the popular leader-follower architecture with the leader robot uses haptic feedback generated by the catheter's interaction with the environment through the navigation system~\cite{molinero2019haptic} and maintains an intuitive control that replicates human motion patterns such as insertion, retraction, and rotation. The follower robot mimics the leader motion, and it is made up of two pneumatic linear motors for translation, one pneumatic rotating stepper motor, and two pneumatic J-clamps for clamping the instrument while performing translational motions. 

% \begin{figure}[h]
% 	\centering
%     \includegraphics[scale=0.68]{figures/cathbot/Cathbot follower (2).pdf}
%     \vspace{1ex}
% 	\caption{The design of Cathbot's follower robot. The two clamps lock the catheter in place by using pneumatic actuators. Those are situated on the main rail that facilitates the translational moves of the catheter. The figure also depicts the two rotary motors responsible for the rotational movements of the catheter and guidewire respectively.}
% 	\vspace{0.3cm}
% 	\label{fig:cathbot}
% \end{figure}

Given the linear mapping between the leader and the follower robot in CathBot's design~\cite{kundrat2021mr}, we focus only on simulating the follower robot for simplicity. %The simulated model of the follower robot has been created through the decomposition of its CAD model into several sub-parts. 
We simulate the follower robot by constructing four modular platforms that are attached to the main rail. On two of those platforms, a pair of clamps is set to secure the guidewire in place during the translational movements, whilst on the other two, rotary catheter and guidewire platforms are attached for performing the angular motions. The parts that account for the translational movements on the main rail as well as the clamps are joined using prismatic joints. Furthermore, revolute joints are used to bind the wheels, thus providing the catheter with rotational movements. The rotational aspect of the catheter implies frictional movement. That is, when the rotational movement is actuated, the clams lock the catheter in place and rotate it. A similar procedure is carried out when the catheter is linearly displaced. However, this friction-reliant rotation is difficult to undertake in simulation, and therefore we assume a perfect motion throughout the system and actuate the joints directly. %To enable real-time simulation, we actuate the catheter directly through the use of prismatic joints. This method retains the control capabilities of the original CathBot design whilst facilitating manipulation learning rather than the underlying physics.

% \subsection{System Overview}

% \begin{figure}[h]
% 	\centering
% 	\includegraphics[width=0.99\linewidth]{figures/CathSim_overview_v2.pdf}
% 	\vspace{2ex}
% 	\caption{TBD. Link to figure: https://drive.google.com/file/d/17vquhVnh0h1q9cpawihRoSM6LcxxYXgR/view?usp=sharing}
% 	\vspace{0.3cm}
% 	\label{fig:overview}
% \end{figure}

We simulate the follower of the CathBot robot together with the aortic arch phantoms and the catheter. We chose to model these elements using MuJoCo's~\cite{mujoco} physics engine, given its stability and computational speed while enabling real-time interactions.

\subsection{Aorta Simulation}

\begin{figure}[!ht]
	\centering
	\subfigure[]{\includegraphics[width=0.39\linewidth]{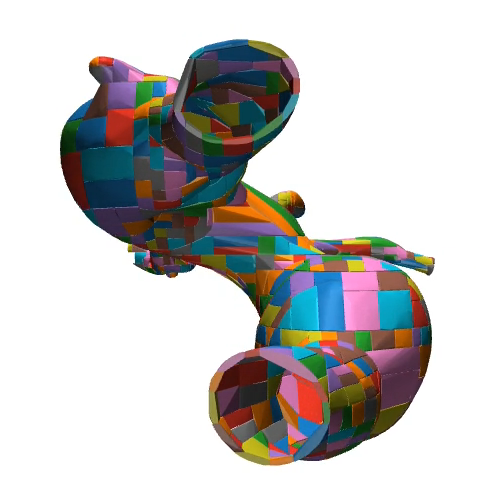}}
	\subfigure[]{\includegraphics[width=0.59\linewidth]{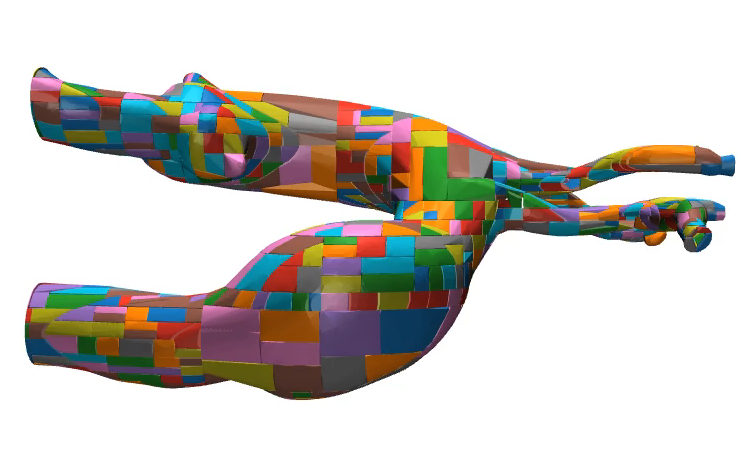}}
		\vspace{1ex}
	\caption{The collision property of the aorta is inferred through the decomposition of the aortic arch into a series of convex hulls, whilst the visual properties are given by scanned high fidelity silicone-based, transparent, anthropomorphic phantoms.}
	\label{fig:aorta}
\end{figure}
To simulate the aorta, we first scan the silicone-based, transparent, anthropomorphic phantom, of the Type-I and Type-II aortic arch model (Elastrat Sarl, Switzerland) to create high-fidelity 3D mesh models. The concave mesh is then decomposed into a set of nearly convex surfaces using the volumetric hierarchical approximate decomposition~\cite{silcowitz2010interactive} resulting in $1,024$ convex hulls for Type-I aortic arch and $470$ convex hulls for Type-II aortic arch. The difference in the number of convex hulls is given by the difference in the measure of concavity of the two meshes~\cite{mamou2009simple}. The convex hull is used to model the collision, as it aids the computational process and allows the use of soft contacts by the physics engine~\cite{mujoco}. We show the aorta simulation in Fig.~\ref{fig:aorta}. % We also include the original mesh without contact in order to maintain the visual realism of the system. 

\subsection{Catheter Simulation} 

\begin{figure}[h]
	\centering
	\includegraphics[scale=0.96]{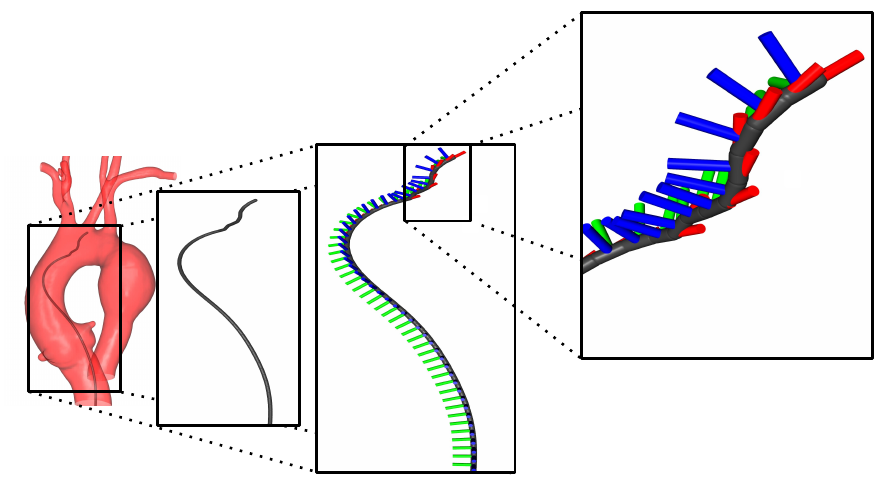}
	\vspace{2ex}
	\caption{The catheter's continuous properties are approximated by a series of rigid bodies linked through revolute joints. The kinematic chain can be visualized through the frames of each individual body, where the last ten links are actuated by internal micro-motors.}
	\vspace{0.3cm}
	\label{fig:catheter}
\end{figure}

One form of modelling the catheter represents the discretization of the continuous shape into a series of rigid bodies, interconnected by revolute or spherical joints~\cite{burgner2015continuum}. This approach has been further categorized into discrete and serpentine methods, where the latter differs by having shorter rigid segments and, implicitly, a higher number of links~\cite{robinson1999continuum}. The discretization approach has been proven to confer reasonable accuracy for shape prediction of the continuum robot~\cite{shi2016shape}. % where the authors paired the robot with Fibber Bragg Gratings (FBG) sensors. 
%% Constant Curvature
In contrast to the discretization approach, the continuous curve estimation methods~\cite{burgner2015continuum} assume that the described object is composed of an elastic backbone% where the model aims to describe the shape of the underlying backbone
. The constant curvature model describes the continuum robot geometry with a finite number of mutually tangent curved segments, each with a constant curvature along its length or a variable curvature approach that takes into account the different curvatures across the backbone of the robot~\cite{burgner2015continuum}.

In this work, to enable real-time performance, we model the catheter based on a discretization approach by creating a serpentine-like model in which its continuous shape and deformation are approximated by a series of rigid bodies and revolute joints. A prismatic joint was added for mimicking the direct linear movement of the catheter. Similarly to~\cite{kwok2012dimensionality}, we limit the actuation to the tip of the catheter, where the revolute joints are intrinsically actuated by motors. This approach allows the use of standardized modelling based on Newton-Euler equation~\cite{featherstone2014rigid} which is a good approximation of the inherent continuous shape that benefits from the computational efficiency of the method. The catheter model is shown in Fig.~\ref{fig:catheter}.

%In this work, to enable real-time performance, we model the catheter based on a discretisation approach, creating a serpentine-like model in which its continuous shape and deformation are approximated by a series of 100 rigid bodies joined together by two revolute joints, resulting in 198 revolute joints. This configuration transfers into a catheter length of 72~cm whereas an endovascular-based catheter has a length of 65-125~cm. A prismatic joint was added for mimicking the direct linear movement of the catheter resulting in an 199 degrees of freedom system. We limit the actuation to the tip of the catheter, represented by the last 10 links or 7~cm, where the revolute joints are intrinsically actuated by motors, similarly to~\cite{kwok2012dimensionality}. This approach allows the use of standardised modelling based on Newton-Euler equations which is a good approximation of the inherent continuous shape that benefits from the computational efficiency of the method. The catheter model is shown in Fig.~\ref{fig:catheter}. 

\subsection{Contact Simulation}
% As our \textbf{CathSim} is based on MuJoCo~\cite{mujoco}, the contact between the aorta and the catheter are simulated using point contacts. In practice, point contact is defined geometrically as a point between two geometric objects and a spatial frame centered at that point in a global coordinate frame. The first axis of this frame is the contact normal direction, while the two other axes define the tangent plane. The contact distance is then used to determine if the penetration happens (i.e, the contact distance is positive if two geometric objects are separated, zero when they are in contact, and negative when they penetrate)~\cite{mujoco}. 
% As our \textbf{CathSim} is based on MuJoCo~\cite{mujoco}, the contact between the aorta and the catheter are simulated using point contacts. In practice, point contact is defined geometrically as a point between two geometric objects and a spatial frame centered at that point in a global coordinate frame. The first axis of this frame is the contact normal direction, while the two other axes define the tangent plane. The contact distance is then used to determine if penetration happens (i.e., the contact distance is positive if two geometric objects are separated, zero when they are in contact, and negative when they penetrate)~\cite{mujoco}. While this approach does present some limitations, such as a simplified representation of real-life contact, it is a reasonable compromise that balances computational efficiency and a sufficient degree of realism. 
As our \textbf{CathSim} is based on MuJoCo~\cite{mujoco}, the contact between the aorta and the catheter are simulated using point contacts. In practice, point contact is defined geometrically as a point between two geometric objects and a spatial frame centered at that point in a global coordinate frame. The first axis of this frame is the contact normal direction, while the two other axes define the tangent plane. The contact distance is then used to determine if penetration happens (i.e., the contact distance is positive if two geometric objects are separated, zero when they are in contact, and negative when they penetrate)~\cite{mujoco}. While this approach does present some limitations, such as a simplified representation of real-life contact, it is a reasonable compromise that balances computational efficiency and a sufficient degree of realism~\cite{todorov2011convex}.

\subsection{How will CathSim be useful to the community?}
We discuss some interesting research directions that can be benefited from our simulator:
\begin{itemize}
    \item Training and benchmarking autonomous catheterization agents: While several works~\cite{behr2019deep,karstensen2020autonomous,kweon2021deep} have attempted to tackle this problem, they do not use the same evaluation system. We hope that \textbf{CathSim} will be used as the benchmarking framework for this task.
    \item Sim-to-Real: With our simulator, we can train agents in simulation and transfer them to real robots~\cite{karstensen2020autonomous}. This approach has been applied in other fields such as autonomous driving and achieved substantial benefits~\cite{chi2020collaborative}.

\end{itemize}
Besides these tasks, we believe our simulator can be used in other scenarios such as developing surgical planning~\cite{li2021path} and medical training~\cite{nesbitt2016role}. The research community is free to explore other applications of our simulator.

\section{Reinforcement Learning for Autonomous Cannulation} \label{Sec:acnn}

We consider the task of autonomous cannulation, where our system represents an episodic Partially Observable Markov Decision Process (POMDP). Our agent, represented by the catheter, interacts with an environment $\mathcal{E}$ represented by an aortic arch type.  At each time step $t$, the agent, receives an observation $s_t$, chooses an action $a_t(s_t)$, receives a reward $r_t(s_t,a_t)$ and arrives in a new state $s_{t+1}$. The episode terminates when the agent reaches the goal position within the aorta $g \in \mathcal{G}$.

\subsection{Observations}

We consider three types of observations for the experiments, namely Internal, Image, and Sequential. In the Internal setup, we fed the system extensive data such as \textit{position, velocity, center of mass inertia and velocity, actuator-generated force, and external forces on the body}. For the Image observation, we utilized a virtual RGB camera positioned above the aortic phantom, producing \(128 \times  128\) resolution grayscale images akin to X-ray images used in clinical procedures. Despite the possibility of higher resolution, our experiments indicated no performance improvement, only increased computational demand and memory requirements.

Moreover, due to the difficulty of inferring the catheter actions given one image as in the “Image” observation setup, we consider the “Sequential” observation by inserting the temporal dimension through the concatenation of three subsequent images  $\{s_{t-2}, s_{t-1}, s_{t}\}$. This observation would take into account the temporal domain of the catheter action and potentially provide more information to the RL agent. 

\subsection{Actions and Rewards}

The actions are represented by a vector $a_t \in \mathbb{R}^{21}$, where the first $20$ elements are associated with motors that actuate the revolute joints inside the tip of the catheter plus the prismatic joint responsible for the translational movement. Furthermore, the actions are normalized within an $[-1,1]$ interval, giving a space of  $a_t \in [-1,1]^{21}$. 

Given the sparsity of the reward function, we chose to convey more spatial information through reward shaping. Considering the navigation task, we provide the agent with an informational reward regarding the distance towards the target by computing the negative Euclidean distance between the head of the catheter $h$ and the goal $g$, such as $d(h,g)=||h-g||$. If the agent is within a distance $\delta$ of the goal, the episode terminates and the agent receives a further reward of $r=10$. The distance of $\delta = 8$~mm was selected as within this distance, the catheter tip is fully inserted within the artery. 

\begin{equation}
    r(h_t, g) = \begin{cases} 
    10      & \text{if} \quad  d(h,g) \leq \delta \\
    -d(h,g) &  \text{otherwise} \\
    \end{cases}
    \label{eq_example}
\end{equation}

% Limitation:local minima
Whilst this reward function assists in agent convergence, it is also prone to local minima. An example would be the erroneous insertion of the catheter in another artery. In this case, the catheter would have to increase the distance from the target to achieve the goal objective. 

% PARAMETERS

% PPO:
%   normalize: true
%   n_envs: 1
%   policy: 'MlpPolicy'
%   n_timesteps: !!float 1e6
%   batch_size: 32
%   n_steps: 512
%   gamma: 0.99
%   learning_rate: 5.05041e-05
%   ent_coef: 0.000585045
%   clip_range: 0.1
%   n_epochs: 20
%   gae_lambda: 0.95
%   max_grad_norm: 1
%   vf_coef: 0.871923

\begin{table*}[hbt!]
\centering
\caption{Results Summary}
\vspace{0.3cm}
\label{tab:1}
\begin{tabular}{@{}llrrrrrrrrrrr@{}}\toprule
\multicolumn{1}{c}{\multirow{2}{*}{Aorta}} & \multicolumn{1}{c}{\multirow{2}{*}{Observation}} &      \multicolumn{2}{c}{Reward} &          \multicolumn{2}{c}{Mean Force~$\mathrm{(N)}~\downarrow$}  &      \multicolumn{2}{c}{Max Force~$\mathrm{(N)}~\downarrow$} &  \multicolumn{2}{c}{Success~\%} \\
\cmidrule{3-4} \cmidrule{5-6} \cmidrule{7-8} \cmidrule{9-10}
            &       & \multicolumn{1}{c}{BCA} & \multicolumn{1}{c}{LCCA} &  \multicolumn{1}{c}{BCA} & \multicolumn{1}{c}{LCCA} & \multicolumn{1}{c}{BCA} & \multicolumn{1}{c}{LCCA} & \multicolumn{1}{c}{BCA} & \multicolumn{1}{c}{LCCA} \\
\midrule
    \multirow{6}{*}{Type-I} 
    & PPO-Internal   &  $-70  \pm 28$  & $-154 \pm 66$ & $0.011 \pm 0.021$ & $0.008 \pm 0.015$ & $0.121 \pm 0.095$ & $0.123 \pm 0.089$ & $53$ & $07$ \\
    & SAC-Internal   &  $-54  \pm 36$  & $-158 \pm 83$ & $0.007 \pm 0.014$ & $0.006 \pm 0.015$ & $0.059 \pm 0.047$ & $0.095 \pm 0.039$ & $77$ & $07$ \\
    &      PPO-Image &  $ -65 \pm 52$  & $\textbf{-140} \pm 96$ & $\textbf{0.005} \pm 0.007$ & $0.007 \pm 0.016$ & $\textbf{0.048} \pm 0.020$ & $0.076 \pm 0.059$ & $83$ & $\textbf{37}$ \\
    &      SAC-Image &  $-391 \pm 157$ & $ -196 \pm 5$ & $0.024 \pm 0.035$ & $\textbf{0.003} \pm 0.003$ & $0.258 \pm 0.094$ & $\textbf{0.059} \pm 0.028$ & $00$ & $00$ \\
    & PPO-Sequential &  $\textbf{-57} \pm 36$   & $-336 \pm 55$ & $0.006 \pm 0.009$ & $0.010 \pm 0.015$ & $0.045 \pm 0.022$ & $0.094 \pm 0.036$ & $\textbf{97}$ & $00$ \\
    & SAC-Sequential &  $-200 \pm 20$  & $-227 \pm 12$ & $0.003 \pm 0.003$ & $0.003 \pm 0.003$ & $0.043 \pm 0.009$ & $0.061 \pm 0.022$ & $00$ & $00$ \\
    \midrule
    \multirow{6}{*}{Type-II} 
    & PPO-Internal   &  $-52 \pm 48$ & $-22 \pm 30$ & $0.016 \pm 0.023$ & $0.011 \pm 0.015$ & $0.118 \pm 0.103$ & $0.068 \pm 0.040$ & $50$ & $87$ \\
    & SAC-Internal   &  $   \textbf{0} \pm 0$ & $-67 \pm 30$ & $0.014 \pm 0.010$ & $0.010 \pm 0.016$ & $\textbf{0.030} \pm 0.007$ & $0.121 \pm 0.044$ & $\textbf{100}$ & $27$ \\
    &      PPO-Image &  $-20 \pm 47$ & $ \textbf{-8} \pm 10$ & $0.017 \pm 0.025$ & $0.015 \pm 0.022$ & $0.046 \pm 0.047$ & $0.066 \pm 0.054$ & $87$ & $\textbf{97}$ \\
    &      SAC-Image &  $-68 \pm 91$ & $-54 \pm 7$  & $\textbf{0.004} \pm 0.005$ & $\textbf{0.005} \pm 0.005$ & $0.042 \pm 0.013$ & $\textbf{0.050} \pm 0.008$ & $73$ & $03$ \\
    & PPO-Sequential & $-47 \pm 49$  & $-80 \pm 92$ & $0.011 \pm 0.017$ & $0.011 \pm 0.018$ & $0.068 \pm 0.059$ & $0.066 \pm 0.050$ & $57$ & $67$ \\
    & SAC-Sequential & $-99 \pm 18$  & $-85 \pm 15$ & $0.005 \pm 0.007$ & $0.006 \pm 0.008$ & $0.055 \pm 0.017$ & $0.100 \pm 0.023$ & $03$ & $00$ \\
\bottomrule
\multicolumn{10}{l}{%\textit{Note: The results were gathered during 30 episodes, each with a different random seed.}
}\\
\end{tabular}
\end{table*}

% Cnn Architecture
% self.cnn = nn.Sequential(
%     nn.Conv2d(n_input_channels, 32, kernel_size=8, stride=4, padding=0),
%     nn.ReLU(),
%     nn.Conv2d(32, 64, kernel_size=4, stride=2, padding=0),
%     nn.ReLU(),
%     nn.Conv2d(64, 64, kernel_size=3, stride=1, padding=0),
%     nn.ReLU(),
%     nn.Flatten(),
% )

\subsection{Network Architectures} 

Considering the continuous action representation, we employ two state-of-the-art RL algorithms, namely PPO~\cite{schulman2017proximal} and SAC~\cite{haarnoja2018soft} with the parameters utilized in \cite{dalal2018safe}. A Multi-Layered Perceptron (MLP) based policy is used for the Internal observation and a Convolutional Neural Network (CNN) is used for the Image and Sequential observation. Note that, for simplicity, we follow~\cite{mnih2013playing} and use CNN for both Image and Sequential observations as they only have one and four input channels respectively where each channel  represents a greyscale image. In our implementation, the MLP has two hidden layers of sizes $64$ with a tanh activation function. The CNN has three convolutional layers with a ReLU activation function. The ADAM optimizer is used to train all networks with a learning rate of $0.0003$.  

\begin{figure}[t]
	\centering
	    \subfigure[Type-I Aortic Arch]{\includegraphics[width=0.49\linewidth]{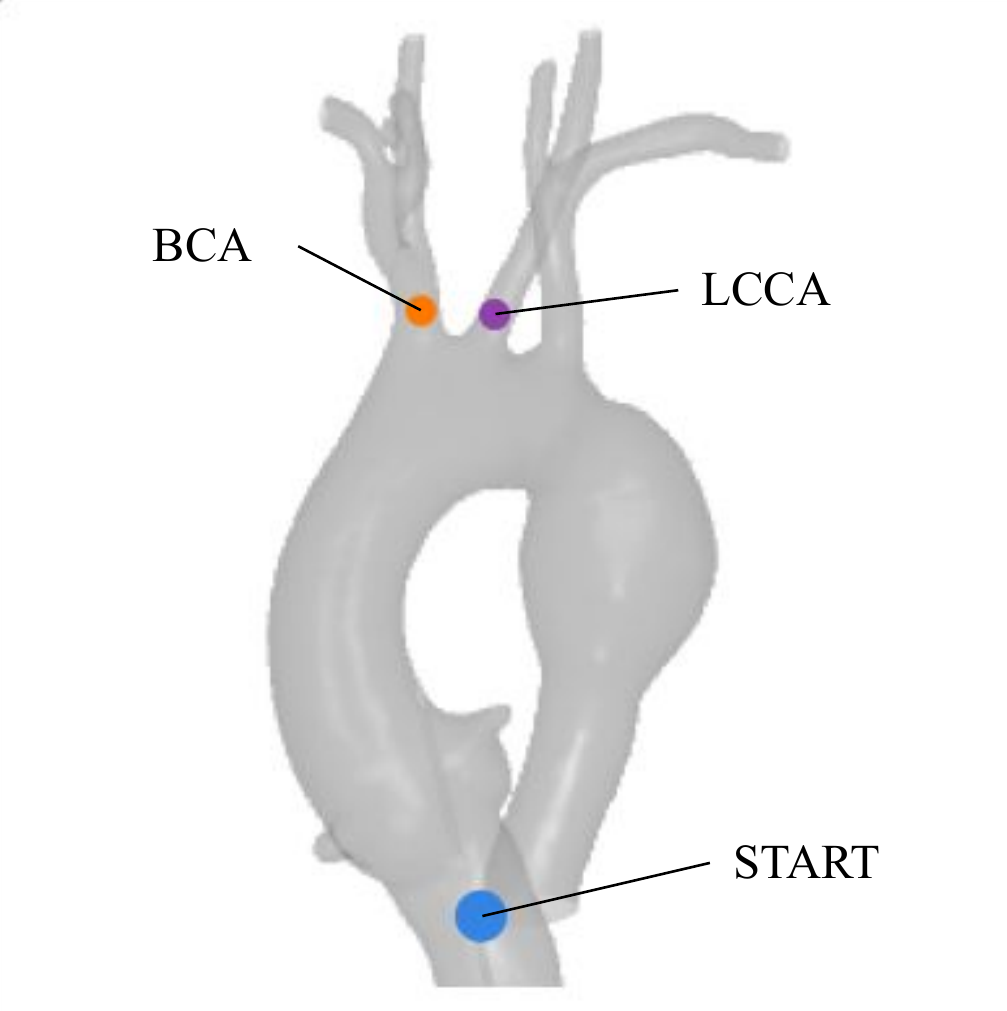}}
    	\subfigure[Type-II Aortic Arch]{\includegraphics[width=0.49\linewidth]{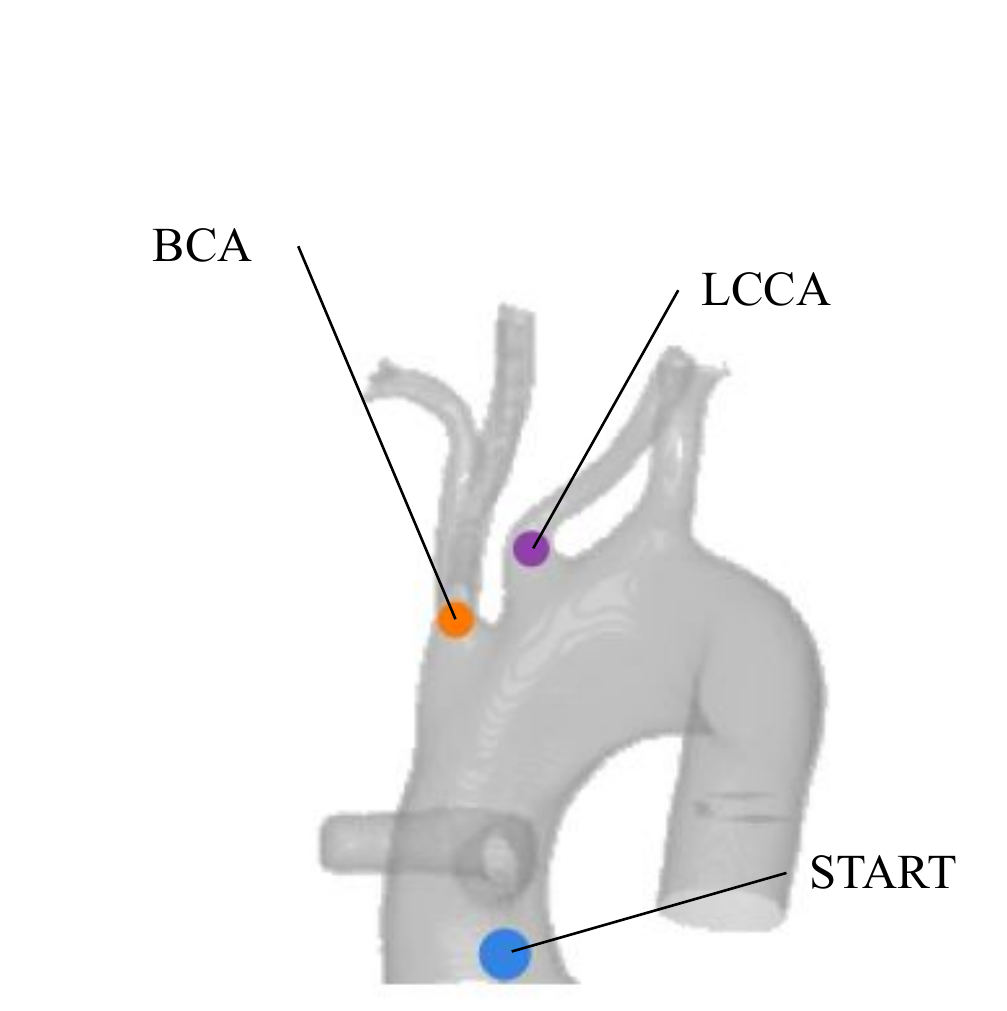}}
    \vspace{1ex}
	\caption{The starting configuration. The figure depicts the navigation task employed in the Type-I and Type-II Aortic Arches. The catheter is initially situated within the ascending aorta with the task of navigating towards the  brachiocephalic artery (BCA) or the left common carotid artery (LCCA). The task finishes when the tip of the catheter is situated within proximity of $8$~mm of the targets.}
	\vspace{0.3cm}
	\label{fig:settup}
\end{figure}

\section{Experiments} \label{Sec:exp}
In this section, we perform intensive experiments to validate our \textbf{CathSim}. We start with the experiment to verify whether our simulator can mimic the behavior of the real-robot (CathBot). We then demonstrate how \textbf{CathSim} can be used for autonomous cannulation tasks using RL algorithms.

\subsection{Simulator Validation}

\begin{figure}[h]
	\centering
	  \includegraphics[scale=1.0]{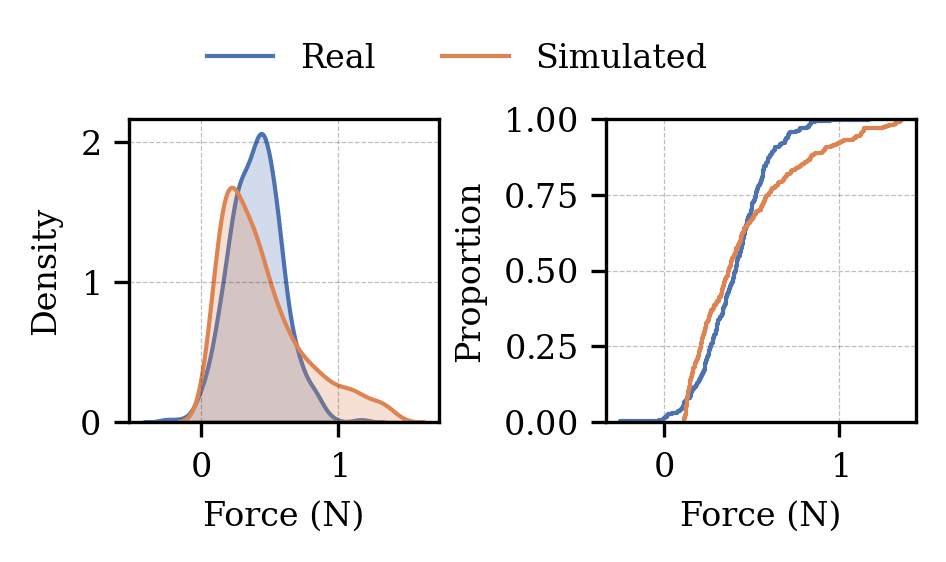}
    \vspace{1ex}
	\caption{The forces distributions are visualized through kernel density estimation (left) as well as through the cumulative distribution function (right). The figure depicts the similarities between the two distributions, where the empirical distribution gathered from the simulator shows a heavier tail than the normal distribution. Within the right subfigure, it can be observed that the cumulative distributions intersect at the $0.5$ threshold, which signifies that the two distributions have a similar median ($\tilde{x_1} \sim \tilde{x_2}$).}
	\vspace{0.3cm}
	\label{fig:force_distributions}
\end{figure}

\subsubsection{Setup}

We assess the validity of our simulator, considering the distribution of forces generated throughout cannulation of the brachiocephalic artery (BCA). Ideally, we want to compare the force measured by our simulator and the force measured in the real-world experiment setup~\cite{kundrat2021mr}. Note that, in the experiments conducted in~\cite{kundrat2021mr}, a load cell (Mini40, ATI Industrial Automation, Apex, NC, USA) was used to capture the force generated by the interaction of the instruments with the silicon phantom. 

To extract the force from our simulator, we manually perform the cannulation using the keyboard, then in each simulation time step, we extract the collision points between the catheter and the aorta along with the tridimensional force expressed as the normal force $f_z$ and frictional forces $f_x$ and $f_z$. We further compute the magnitude of the force given the previous, such as given a time instance $t$, the magnitude of the force $f_t$ is given by

\begin{equation}
f_t = \sqrt{ f_{x,t}^2 + f_{y,t}^2 + f_{z,t}^2}  \label{eq:2}
\end{equation}

We proceed by offering a comparison between the observed empirical distribution and a normal distribution derived from the real experiments conducted in~\cite{kundrat2021mr}.
%during the cannulation of the right common carotid artery~(RCCA). In the experiments previously mentioned, a load cell is situated underneath the phantom, which captures the force inflicted by the catheter on the aorta. The target distribution has been chosen to be in the proximity of brachiocephalic artery.  As such, 
%We assume that the force distribution of both experiments follows the normal distribution ${X}(\mu, \sigma)$. % with the mean of $\mu=0.4$ and a standard deviation of $\sigma = 0.02$, from which 
We sample the same number of samples generated in our simulation and the real experiment in~\cite{kundrat2021mr} for a fair comparison. We use the distributions to generate a cumulative distribution function $F_X(x)=P(X \leq x)$, which can be visualized in Fig.~\ref{fig:force_distributions}. We further assess the normality of the sampled data, followed by a comparison of the two given distributions.

\subsubsection{Results} 

We begin our comparison with the Shapiro-Wilk test of normality on the data extracted from the simulator and, given a p-value of $p \approx 7.195\times 10^{-17}$ and a statistic of $0.878$, we conclude that the sampled data does not represent a normal distribution ${X}(\mu, \sigma)$. This can be visualized in Fig.~\ref{fig:force_distributions}, where the sampled data is plotted against a normal distribution. Furthermore, we assess the homoscedasticity of the sample distribution and normal distribution by using Levene's tests, which results in a statistic of $40.818$ and a p-value of $p\approx 2.898\times 10^{-10}$, therefore, concluding that $\sigma_1^2 \neq \sigma_2^2$. Given the previous statistics (i.e., non-normal distribution and unequal variances), we select the  non-parametric Mann-Whitney test to compare the given distributions. The resulting statistic given the test is $76076$, with a p-value of $p\approx0.445$. Given that the p-value is higher than the threshold of $p=0.05$, we can conclude that the differences in the distributions are merely given to chance and therefore the distributions can be considered as being part of the same population and thus convene that the force distribution of our simulator closely represents the distribution of forces encountered in the real-life system. Therefore, we can see that our \textbf{CathSim} successfully mimics the behavior of the real-world system.

\subsection{Reinforcement Learning Results}

% \textbf{In general, I recommend reducing the text "Quantitative Results" and "Qualitative Results". This result is not strong and there is too much unnecessary discussion}

% \textbf{Do not duplicate or repeat simple points. For example, Fig 6 is mentioned several times, and not clear why you need to do that. Just explain it one time, clearly.  }

\subsubsection{Setups} We consider the autonomous catheterization of two principal arteries into the aortic arch, namely the brachiocephalic artery (BCA) and the left common carotid artery (LCCA). Within both setups, we position the catheter tip at the starting locations within the ascending aorta and terminate the training when the catheter is fully inserted within the artery. We follow the same procedure for both the Type-I and the Type-II aortic arches. Please see Fig.~\ref{fig:settup} for our experimental setup.

We trained the model for $600,000$ time steps. Each episode started with a random catheter displacement of $1$~mm, followed by navigation through the ascending aorta to reach the goal. If not achieved within $2,000$ steps, the episode ended. Data on the aorta's contact points and exerted force, as per Eq.~\ref{eq:2}, was gathered at each step to form a force heatmap overlaid on the RGB virtual image. The model's performance was evaluated over $30$ episodes. The maximum and mean forces for $n$ samples were computed as:
%$\max\limits_{t \in n} f_t$ and 

\begin{equation}
    f_{\rm max} = \max\limits_{t \in n} f_t, \quad \quad 
    f_{\rm mean}=\frac{1}{n}\sum\limits_{t=1}^n f_t
\end{equation}

The training time ranged from $1$ hour for PPO to $8$ hours for SAC, making PPO approximately five times faster. Regardless, our simulator showed a performance of $60$ FPS for the image-based environment and $80$ FPS for the internal-based environment.

\begin{figure}[h]
	\centering
	    \subfigure[Type-I Aortic Arch - BCA]{\includegraphics[width=0.45\linewidth]{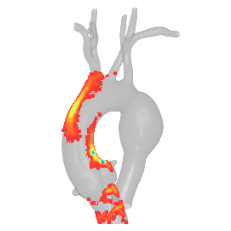}}\hspace{1ex}
    	\subfigure[Type-II Aortic Arch - BCA]{\includegraphics[width=0.45\linewidth]{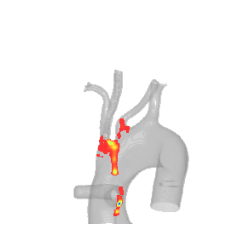}}\hspace{1ex}
        \subfigure[Type-I Aortic Arch - LCCA]{\includegraphics[width=0.45\linewidth]{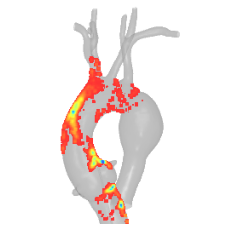}}\hspace{1ex}
    	\subfigure[Type-II Aortic Arch - LCCA]{\includegraphics[width=0.45\linewidth]{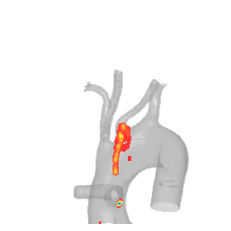}}\hspace{1ex}
    	\vspace{3ex}
	\caption{The force heatmap extracted from the interaction between the catheter and the aortas when undertaking the navigating towards the BCA target on the top row, and LCCA target on the second row. The catheterization procedure exerts a greater force on the first aorta where complicated maneuvers are required to reach the targets, whilst on the second aorta it generates less force.}
	\label{fig:heatmap}
\end{figure}

\subsubsection{Quantitative Results}

% Using the extracted forces for each time step, we compute the mean force of the in-contact zones, as well as the maximum force exerted on the aorta during an episode. We then compute the average across the $30$ episodes for both the maximum and the mean force. Furthermore, we compute the mean reward over the evaluated episodes and display the results in 
Table~\ref{tab:1} shows the force, reward, and success rate of all methods. Type-I Aortic Arch experiments show that PPO relying on a sequential observation space achieved the greatest reward when cannulating the BCA target, although it shows the least performance when the target is LCCA. A more coherent reward has been achieved while using a singular image observation, where the cannulation of LCCA presents the greatest success~($34\%$). In contrast, the cannulation of BCA presents close results to the sequential observation~($83\%$). The performance gap between the cannulation of BCA and LCCA is mainly because of the start configuration and the position of the BCA and LCCA in the aortic arch. Naturally, from Fig.~\ref{fig:settup} we can see that it would be easier for both humans and RL agents to reach BCA rather than LCCA since LCCA's position is further away from the navigation direction and is surrounded by other vascular branches. %. and One potential explanation for this divergence in outcomes could be the inherent randomness that is present within reinforcement learning (RL) methods, which may lead to varying performances on different goal configurations. 

% Type-II aortic arch
Given the success rates of the cannulation, the catheterization of the Type-II aortic arch appears to be a more straightforward procedure. The task implies that the dependability of the vessel wall for maneuvering is reduced, leading to a more straightforward catheterization procedure. This phenomenon can be also observed in Fig.~\ref{fig:heatmap}, where the catheter exerted a greater amount of force on the aortic walls in order to reach the designated target. Overall, from our experiments, PPO shows better suitability for the task, especially when the observation is image-based.

\begin{figure}[t]
	\centering
	\includegraphics[scale=1]{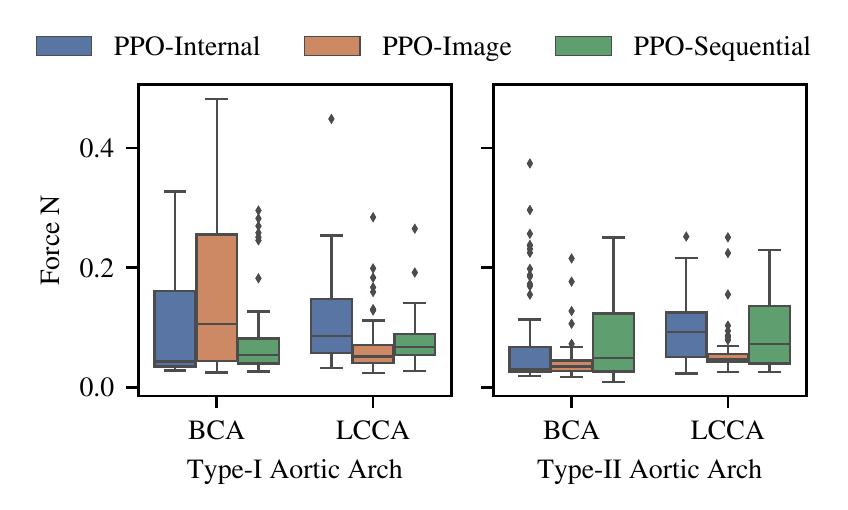}
    \vspace{1ex}
	\caption{The max forces generated by the catheter on the aorta within each of the evaluation episodes during the PPO training are shown. The left figure depicts the results gathered within the Type-I Aortic Arch where the catheter has greater interaction with the aorta, whereas the one on the right shows the results gathered within the Type-II Aortic Arch where the navigation is more direct.}
	\vspace{0.3cm}
	\label{fig:boxplot}
\end{figure}

\subsubsection{Qualitative Result}

% Utilizing force frames from each time step, we calculate the mean of in-contact regions across all episodes. This data is superimposed on the phantom model via the simulator's projection matrix to create force heatmaps for all target configurations, as displayed in Fig.~\ref{fig:heatmap}. These maps reveal greater force application by the catheter at bend points needed to reach targets. For instance, in Fig.~\ref{fig:heatmap}~(c), the catheter bends on the ascending aorta to reach the brachiocephalic and left common carotid artery, often overshooting the LCCA target and deviating towards the descending aorta. In the Type-II Aortic Arch, the catheter exerts more force on the wall between BCA and LCCA, especially towards BCA. In comparison, the Type-I Aortic Arch presents more challenges.
Force frames from each time step help us calculate the mean of in-contact regions, which are then superimposed on the phantom model to generate force heatmaps (see Fig.~\ref{fig:heatmap}). These maps show increased catheter force at bends required for target access. For example, the catheter often overshoots the LCCA target in Fig.~\ref{fig:heatmap}~(c) whilst exerting greater force on the ascending aortic walls within Type-I Aortic Arch. Within the Type-II Aortic Arch, more force exerted on the wall between BCA and LCCA. Regardless, the Type-I Aortic Arch, poses more difficulties.

We further display the results of the force interaction in Fig.~\ref{fig:boxplot}. Analyzing the figure, it can be seen that the force distribution is quite similar along both the Type-I and Type-II Aortic Arch, where most of the force is concentrated within the $0.0$ to $0.2\mathrm{N}$ range. Despite this, there are outliers which exert a force greater than $0.4\mathrm{N}$. Specifically, the BCA within the Type-I Aortic Arch has a notably wider interquartile range than the LCCA, suggesting that the BCA generally exerts more force than the LCCA.

\begin{figure}[t]
	\centering
	\includegraphics[scale=1.0]{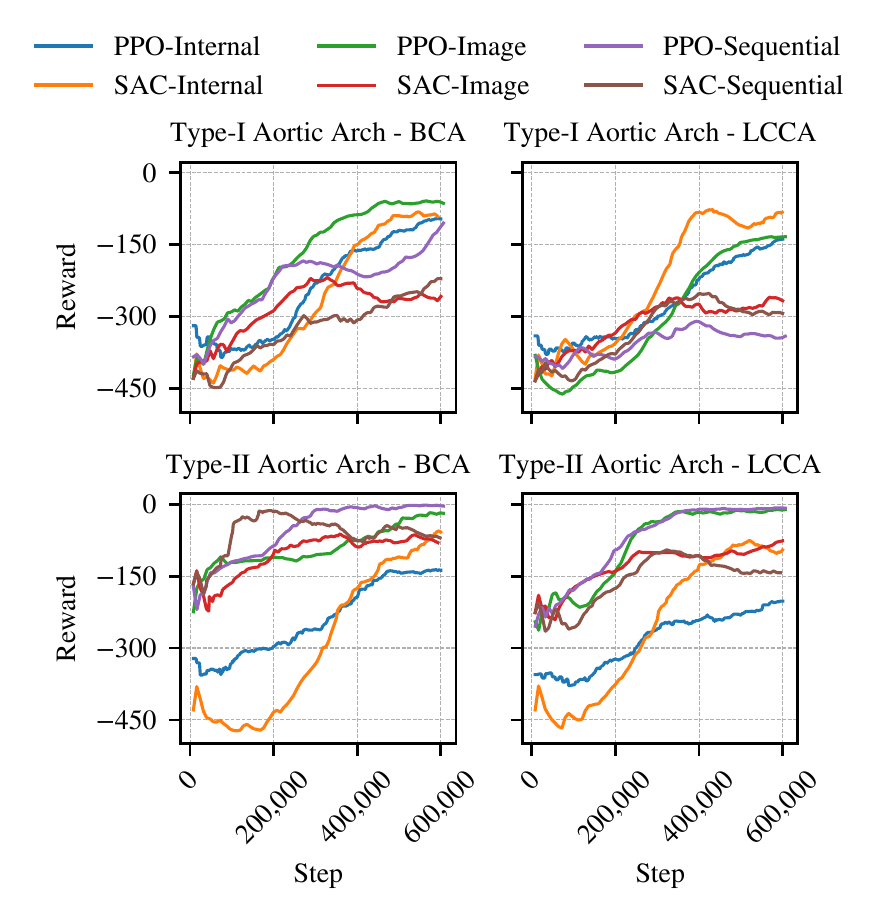}
    %\vspace{3ex}
% 	\includegraphics[scale=.6]{figures/heatmap_1.png}
	\caption{The rewards when training autonomous agents using an on-policy algorithm (PPO) and an off-policy algorithm (SAC). In general, we observe that PPO performs better than SAC. The image observation consistently achieved good results when using PPO. Furthermore, the sequential observation shows the least amount of progress compared with the other observation spaces.}
% 	\textbf{Data:} \url{https://drive.google.com/drive/folders/1wWIlztSqw61F6tdT8G8aRFD0qh2I4OBU?usp=sharing}}
	\label{fig:rewards}
\end{figure}

% At each time step, we compute the mean reward of the last $100$ steps and display the results in Fig.~\ref{fig:rewards}. The algorithms show continuous learning with slight convergence. Within all the environments, PPO with an image observation obtained the highest or close to the highest reward, thus showing that an on-policy algorithm, generally performs better for the given task. Furthermore, it shows that an image-based observation that correlates to the medical procedure undertaken in real clinical scenarios is capable of reaching the most reward. Whilst using the internal observation space, the off-policy algorithm (i.e. SAC) performed reasonably well, managing to obtain the highest reward within the Type-I aortic arch when cannulating the LCCA target whilst obtaining competitive results within the other tasks. Within all cases and coherent with the evaluation undertaken in the previous section, the sequential observation space yielded the least reward.  
At each time step, we compute the mean reward of the last $100$ steps and display the results in Fig.~\ref{fig:rewards}. The algorithms show continuous learning with slight convergence. In our experiments, PPO with an image observation obtained the highest or close to the highest reward. Furthermore, it shows that an image-based observation that correlates to the medical procedure undertaken in real clinical scenarios is capable of reaching the most reward. While using the internal observation space, SAC performed reasonably well, managing to obtain the highest reward within the Type-I aortic arch when cannulating the LCCA target. Within all cases and coherent with the evaluation undertaken in the previous section, the sequential observation space yielded the least reward.

\section{Discussion}
%\textbf{rewrite this - see the "Discussion" from https://arxiv.org/pdf/2303.02401.pdf }
While our current framework represents promising progress, it remains challenged by the complexity of fully mimicking real-world dynamics within a simulated environment. There exist critical gaps in simulating intricate biological structures like a deformable aorta, implementing realistic guidewire simulation, and factoring in the nuanced interplay between surgical instruments and fluid mechanics. Our systematic experimentation, however, has shed light on crucial directions for further advancement. These include: \textit{i)} the development of comprehensive and sophisticated features that aim to lessen the divergence between real-world conditions and simulated environments, \textit{ii)} the integration of our simulation system into Augmented Reality/Virtual Reality (AR/VR) platforms to foster a more immersive and interactive experience, and \textit{iii)} the execution of robust comparative studies that pit the simulated environment against its physical counterparts to validate the simulation's accuracy and reliability. Despite the inherent challenges in achieving a perfect simulation of real-world medical procedures, our research has illuminated promising pathways for subsequent development. We aim to refine our system into a more robust, comprehensive, and precise research tool. Furthermore, to promote collaborative scientific growth, we intend to disseminate our source code to stimulate additional research and innovation in this rapidly evolving field.

\section{Conclusions}\label{Sec:con}
We present \textbf{CathSim}, an open-source simulation environment designed as a comprehensive benchmarking platform for autonomous endovascular navigation. Our proposed simulator allows for the development and testing of a diverse range of algorithms for autonomous cannulation, eliminating the need for physical robotic systems and thereby reducing the associated risks and costs of experimental hardware. It serves a dual purpose as a development and benchmarking platform for computer scientists and roboticists, as well as a training platform for healthcare professionals. Furthermore, with the rapid advancements in medical robotics, \textbf{CathSim} provides a risk-free, controlled environment where innovative procedures and techniques can be tested and refined before their implementation in real-world clinical scenarios. By maintaining an open-source model, our simulator also encourages collaborative advancements and knowledge sharing within the scientific community.

\bibliographystyle{class/IEEEtran}
\bibliography{class/IEEEabrv,class/reference}

\end{document}